\documentclass{article}
\usepackage{iclr2026_conference,times}
\usepackage[T1]{fontenc}
\usepackage[utf8]{inputenc}
\usepackage{amsmath,amssymb,mathtools,booktabs,array,tabularx,multirow}
\usepackage{graphicx,subcaption,xcolor,tikz,listings,float,placeins}
\usepackage{longtable}
\usetikzlibrary{arrows.meta,positioning,fit,calc,backgrounds}
\usepackage{microtype,etoolbox}
\patchcmd{\abstract}{\vspace{0.5ex}}{\vspace{-0.5ex}}{}{}
\usepackage[hyphens]{url}
\usepackage[hidelinks]{hyperref}
\usepackage{cleveref}
\crefname{figure}{Figure}{Figures}
\Crefname{figure}{Figure}{Figures}
\crefname{table}{Table}{Tables}
\Crefname{table}{Table}{Tables}
\crefname{appendix}{Appendix}{Appendices}
\Crefname{appendix}{Appendix}{Appendices}
\crefname{subappendix}{Appendix}{Appendices}
\Crefname{subappendix}{Appendix}{Appendices}
\crefname{subsubappendix}{Appendix}{Appendices}
\Crefname{subsubappendix}{Appendix}{Appendices}
\crefname{subsubsubappendix}{Appendix}{Appendices}
\Crefname{subsubsubappendix}{Appendix}{Appendices}

\usepackage{xspace}

\newcommand{\methodname}{\textsc{How to Train Your Little Dragon}\xspace}
\newcommand{\method}{\methodname}
\newcommand{\TSEL}{\textsc{TSEL}}

\newcommand{\clip}{\operatorname{clip}}

\newcommand{\code}[1]{\texttt{#1}}
\usepackage{comment}
\newcolumntype{Y}{>{\raggedright\arraybackslash}X}
\title{How to Train Your Little Dragon:\\Teaching a Small Language Model to Construct and Refine GPU Optimization Scripts}
\author{%
  \parbox[t]{\dimexpr\textwidth-2\tabcolsep\relax}{%
    \raggedright
    \mbox{Annabelle Sujun Tang$^{1, *}$}, \mbox{Wei Tang$^{2}$},
    \mbox{Songyao Jiang$^{2}$}, \mbox{Tianhui Li$^{2}$},
    \mbox{Shengzhe Xu$^{2}$}, \mbox{Jiin Woo$^{2}$},
    \mbox{Kaan Ozkara$^{2}$}, \mbox{Youngsuk Park$^{2}$},
    \mbox{Zhen Jia$^{2}$}, \mbox{Hadi Esmaeilzadeh$^{1}$}%
  }%
  \\
  $^{1}$University of California, San Diego\\
$^{2}$Amazon Annapurna Labs\\
  \parbox[t]{\dimexpr\textwidth-2\tabcolsep\relax}{%
    \raggedright
    \texttt{\mbox{sujun@ucsd.edu},
      \{weittang, songyaoj, tianhl, shengzx,\\
      woojiin, kaanozka, pyoungsu, zhej\}@amazon.com,
      \mbox{hadi@ucsd.edu}}%
  }%
}
\iclrfinalcopy
\begin{document}
\maketitle
\fancyhead{}
\fancyhead[C]{How to Train Your Little Dragon}

\vspace{-4pt}
\begin{abstract}
Compiler transformations interact, making GPU kernel optimization a search for combinations that preserve the computation while improving performance.
We investigate how to train a small language model to generate the optimization: a correct, performance-improving script that coordinates interacting transformations in one complete proposal.
We introduce \methodname, a training method that teaches a small language model both to construct complete optimization scripts and to improve existing ones.
Rather than learn only to rank search candidates or generate replacement kernel code, our \textsc{Little Dragon} learns to generate correct scripts that transform the under-optimized kernel and improve its performance.
We devise a compact, lossless representation whose deterministic decoding reconstructs the exact compiler trace, preserving its transformation parameters and dependencies.
We use a compiler framework TVM MetaSchedule to build training data on 54,116 distinct workloads, and construct 108,594 examples in two training sets: 60,236 teach the model to generate an optimized script from an unoptimized kernel, while 48,358 teach it to produce a better script given the kernel and a somewhat optimized script.
Supervised fine-tuning teaches the model to construct and improve optimization scripts from these search-derived examples.
We then add reinforcement learning with a shaped reward that combines correctness with measured performance, awarding a latency-based bonus only to numerically correct kernels.
This enables the model learn not only from existing optimizations, but from the correctness and performance of its generated scripts.
We apply this training method to a model with 8 billion parameters.
After supervised fine-tuning and reinforcement learning, the model achieves per-attempt correctness of 91.27\% for construction and 92.01\% for refinement, while reinforcement learning produces faster kernels than supervised training on 88 of 102 construction tasks and 145 of 170 refinement prompts.
With only 8 billion parameters, our \textsc{Little Dragon} achieves higher per-attempt correctness and better kernel performance than six frontier LLMs---GPT-6 Astra,
Opus~5, Kimi-K3, and GLM-5.3 among them, each with hundreds of billions to trillions of parameters---for which task-specific training is
unavailable or expensive.
These results suggest that \methodname enables a small language model to learn from compiler search and direct correctness and performance feedback, internalizing an optimization policy that constructs and improves complete scripts in one shot. \def\thefootnote{*}\footnotetext{Work done during an internship at Amazon Annapurna Labs.}
\end{abstract}

\section{Introduction}
\vspace{-4pt}
The cost of serving a model depends not only on the model itself, but on how efficiently its kernels use the GPU.
Compiler optimization can reduce this cost by turning the same computation into a faster implementation, yet finding that implementation can itself require substantial compilation and measurement.
The difficulty is that transformations interact: tiling, memory placement, and thread mapping must work together to preserve correctness and improve performance.
Search-based compilers discover effective combinations through trial and error, repeatedly generating and evaluating candidates~\citep{zheng2020ansor,shao2022metaschedule}.
We ask how to train a small language model to construct and improve complete optimization scripts in one shot, learning to coordinate interacting compiler transformations into correct, faster kernels.

Existing approaches use learning at different points in optimization.
Learned cost models rank candidates for evolutionary search, while frozen language models
guide exploration without updating their
parameters~\citep{chen2018learning,tang2025reasoningcompiler,tang2026litecoop}.
Other approaches train models to generate scheduling decisions step by step, conditioning each choice on earlier decisions while its benefit may depend on transformations yet to be selected~\citep{zhai2024enabling}.
We set out to explore a research question: can a small language model learn an optimization policy across different tasks that enables it to construct and improve complete optimization scripts in one shot, and how can we train it to do so?

Our hypothesis is that learning to improve an optimization requires examples not only of effective solutions, but of how less effective solutions can become better.
We therefore introduce \methodname, a training method built around two complementary sets of examples.
In the first, we give the model an unoptimized kernel and teach it to generate an optimized script.
In the second, we give it the unoptimized kernel together with a somewhat optimized script and teach it to generate a more optimized script.
Both tasks produce a complete script, allowing refinement to revise earlier and later transformation choices together rather than merely extend the existing sequence.
Through examples from different optimization tasks, we seek to teach the model which combinations work well together and how to improve them, internalizing an optimization policy that it can apply to other kernels.

The supervised examples teach correct, performance-improving optimizations.
We then use reinforcement learning to strengthen the policy through
feedback on the correctness and measured performance of its own complete proposals.
A staged reward recognizes progress toward executable, correct kernels and awards latency bonuses only to numerically correct candidates, with refinement bonuses additionally requiring
improvement over the supplied script.

Both stages require the model to express its complete optimization precisely enough for the compiler to apply it.
A script must specify not only which transformations to use, but how they refer to the loops and memory structures created by earlier transformations.
Raw compiler traces preserve this information through detailed parameters and temporary identifiers, while prose descriptions can explain an optimization without specifying everything needed to reproduce it.
We therefore devise a compact, lossless representation that groups related transformations into typed edits while retaining their order, parameters, and dependencies.
Deterministic decoding reconstructs the exact canonical compiler trace, and round-trip checks verify that representing a source script does not change it.
The model is taught explicit compiler optimization decisions connected to their supporting program evidence. Optimization graphs link compiler-derived facts, optimization analysis, executable schedule edits, and expected effects, exposing both how an optimization is applied and why it is proposed. Search-derived traces fix the executable targets, while static checks ground the accompanying explanations in compiler artifacts. This provides a structured teaching interface for learning compiler optimization through SFT and execution feedback.

To supply examples in this representation, we use TVM MetaSchedule, a compiler framework that explores candidate optimizations through stochastic evolutionary search guided by performance cost models~\citep{shao2022metaschedule}.
From this search, we select the scripts that produce the best-performing validated kernels as training targets.
We pair each target with the unoptimized kernel to teach construction, and additionally supply a somewhat optimized script to teach refinement.
For refinement, we require the target's measured improvement to exceed timing variability, so the example reflects a genuine performance gain.
Across 54,116 distinct workloads, we construct 108,594 training examples: 60,236 for construction and 48,358 for refinement.
Thus, we turn the search process into a source of both correct optimization scripts and concrete examples of how to improve them.

We evaluate whether this training enables a small model to generate correct optimization scripts and improve the performance of the resulting kernels.
We evaluate a model with 8 billion parameters on NVIDIA H100.
After supervised fine-tuning and reinforcement learning, the model achieves per-attempt correctness (Correct@1) of 91.27\% for construction and 92.01\% for refinement. Comparing each model's fixed first output, reinforcement learning produces faster kernels than supervised training on 88 of 102 construction tasks and 145 of 170 refinement prompts for which both outputs are correct and timed.
With only 8 billion parameters, \methodname achieves higher per-attempt correctness and better kernel performance than six frontier LLMs---GPT-6 Astra,
Opus~5, Kimi-K3, and GLM-5.3 among them---for which task-specific training would be unavailable or expensive.
These results provide an affirmative answer to our research question and a concrete way to achieve it: training on construction and refinement examples, followed by rewards for correctness and performance, enables a small model to internalize an optimization policy for generating complete scripts in one shot.

\section{Learning executable compiler transformations}
\label{sec:method}
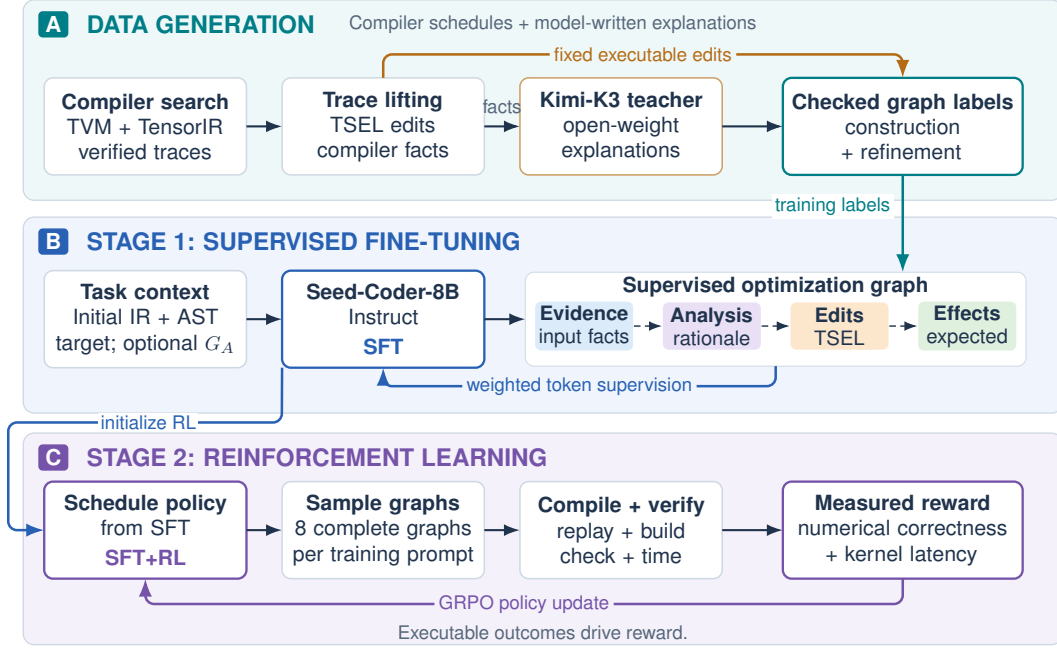
\begin{figure}[t]
\centering
\resizebox{\linewidth}{!}{
\begingroup%
\definecolor{vsInk}{HTML}{20334D}%
\definecolor{vsMuted}{HTML}{56677B}%
\definecolor{vsRule}{HTML}{CFD9E3}%
\definecolor{vsTeal}{HTML}{007E87}%
\definecolor{vsTealBg}{HTML}{EFF8F8}%
\definecolor{vsBlue}{HTML}{2861B3}%
\definecolor{vsBlueBg}{HTML}{F0F5FD}%
\definecolor{vsPurple}{HTML}{7453A8}%
\definecolor{vsPurpleBg}{HTML}{F5F1FA}%
\definecolor{vsAmber}{HTML}{B46B16}%
\definecolor{vsEvidence}{HTML}{DCEBFA}%
\definecolor{vsAnalysis}{HTML}{EAE0F6}%
\definecolor{vsEdits}{HTML}{FCE7CD}%
\definecolor{vsEffects}{HTML}{E0F0DE}%
\begin{tikzpicture}[
  x=1cm,y=1cm,
  font=\sffamily\fontsize{9.5}{11.5}\selectfont,
  text=vsInk,>=Latex,
  panel/.style={rounded corners=5pt,draw=vsRule,line width=.6pt},
  card/.style={rounded corners=4pt,draw=vsRule,fill=white,line width=.7pt,
    align=center,text width=2.95cm,minimum height=1.55cm,inner sep=4pt,outer sep=0pt},
  flow/.style={->,draw=vsInk,line width=.95pt},
  feedback/.style={->,line width=1.15pt,rounded corners=4pt},
  role/.style={rounded corners=3pt,draw=white,line width=.7pt,
    align=center,text width=1.45cm,minimum height=.78cm,inner sep=2pt,
    font=\sffamily\fontsize{9}{10.5}\selectfont,outer sep=0pt},
  annotation/.style={font=\sffamily\fontsize{8.5}{10}\selectfont,text=vsMuted},
  stage/.style={font=\sffamily\bfseries\fontsize{11}{13}\selectfont,anchor=west},
  badge/.style={rounded corners=2pt,text=white,minimum width=.48cm,
    minimum height=.42cm,inner sep=0pt,font=\sffamily\bfseries\fontsize{10}{11}\selectfont}
]
\hyphenpenalty=10000\exhyphenpenalty=10000
\path[panel,fill=vsTealBg] (0,7.1) rectangle (16.6,10.3);
\node[badge,fill=vsTeal] at (.48,9.91) {A};
\node[stage,text=vsTeal] at (.89,9.91) {DATA GENERATION};
\node[annotation,anchor=west] at (5.08,9.91) {Compiler schedules + model-written explanations};

\node[card] (search) at (1.95,8.28) {
  \textbf{Compiler search}\\
  TVM + TensorIR\\
  verified traces};
\node[card] (lift) at (5.75,8.28) {
  \textbf{Trace lifting}\\
  TSEL edits\\
  compiler facts};
\node[card,draw=vsAmber!65] (teacher) at (9.55,8.28) {
  \textbf{Kimi-K3 teacher}\\
  open-weight\\
  explanations};
\node[card,text width=3.55cm,draw=vsTeal,line width=1.05pt] (corpus) at (14.05,8.28) {
  \textbf{Checked graph labels}\\
  construction + refinement};
\draw[flow] (search.east) -- (lift.west);
\draw[flow] (lift.east) -- node[annotation,above=2pt] {facts} (teacher.west);
\draw[flow] (teacher.east) -- (corpus.west);
\draw[feedback,draw=vsAmber] (lift.north) -- (5.75,9.43)
  -- node[annotation,fill=vsTealBg,text=vsAmber,inner sep=2pt] {fixed executable edits}
  (14.05,9.43) -- (corpus.north);

\path[panel,fill=vsBlueBg] (0,3.7) rectangle (16.6,6.83);
\node[badge,fill=vsBlue] at (.48,6.44) {B};
\node[stage,text=vsBlue] at (.89,6.44) {STAGE 1: SUPERVISED FINE-TUNING};
\node[card] (context) at (1.95,5.2) {
  \textbf{Task context}\\
  Initial IR + AST\\
  target; optional $G_A$};
\node[card,draw=vsBlue,line width=1.05pt] (sft) at (5.75,5.2) {
  \textbf{Seed-Coder-8B}\\
  Instruct\\[2pt]
  \textcolor{vsBlue}{\textbf{SFT}}};
\node[rounded corners=4pt,draw=vsRule,fill=white,minimum width=8.0cm,
  minimum height=1.5cm,inner sep=0pt,outer sep=0pt] (graph) at (12.02,5.2) {};
\node[font=\sffamily\bfseries\fontsize{9.5}{11}\selectfont] at (12.02,5.72)
  {Supervised optimization graph};
\node[role,fill=vsEvidence] (evidence) at (8.96,5.09) {\textbf{Evidence}\\input facts};
\node[role,fill=vsAnalysis] (analysis) at (11.0,5.09) {\textbf{Analysis}\\rationale};
\node[role,fill=vsEdits] (edits) at (13.04,5.09) {\textbf{Edits}\\TSEL};
\node[role,fill=vsEffects] (effects) at (15.08,5.09) {\textbf{Effects}\\expected};
\draw[flow,densely dashed,line width=.65pt] (evidence.east) -- (analysis.west);
\draw[flow,densely dashed,line width=.65pt] (analysis.east) -- (edits.west);
\draw[flow,densely dashed,line width=.65pt] (edits.east) -- (effects.west);
\draw[flow] (context.east) -- (sft.west);
\draw[flow] (sft.east) -- (graph.west);
\draw[feedback,draw=vsTeal] (corpus.south) -- (14.05,5.95);
\node[annotation,fill=white,inner sep=1.5pt,anchor=east,text=vsTeal]
  at (13.91,7.0) {training labels};
\draw[feedback,draw=vsBlue] (graph.south) -- (12.02,4.13)
  -- node[annotation,fill=vsBlueBg,inner sep=2pt,text=vsBlue] {weighted token supervision}
  (5.75,4.13) -- (sft.south);

\path[panel,fill=vsPurpleBg] (0,0) rectangle (16.6,3.38);
\node[badge,fill=vsPurple] at (.48,2.99) {C};
\node[stage,text=vsPurple] at (.89,2.99) {STAGE 2: REINFORCEMENT LEARNING};
\node[card,draw=vsPurple,line width=1.05pt] (policy) at (1.95,1.83) {
  \textbf{Schedule policy}\\
  from SFT\\[2pt]
  \textcolor{vsPurple}{\textbf{SFT+RL}}};
\node[card] (samples) at (5.75,1.83) {
  \textbf{Sample graphs}\\
  8 complete graphs\\
  per training prompt};
\node[card] (verifier) at (9.55,1.83) {
  \textbf{Compile + verify}\\
  replay + build\\
  check + time};
\node[card,text width=3.55cm,draw=vsPurple,line width=1.05pt] (reward) at (14.05,1.83) {
  \textbf{Measured reward}\\
  numerical correctness\\
  + kernel latency};
\draw[flow] (policy.east) -- (samples.west);
\draw[flow] (samples.east) -- (verifier.west);
\draw[flow] (verifier.east) -- (reward.west);
\draw[feedback,draw=vsPurple] (reward.south) -- (14.05,.65)
  -- node[annotation,fill=vsPurpleBg,inner sep=2pt,text=vsPurple] {GRPO policy update}
  (1.95,.65) -- (policy.south);
\node[annotation] at (8.3,.2)
  {Executable outcomes drive reward.};
\draw[feedback,draw=vsBlue] (sft.south west) -- (4.13,3.55)
  -- (-.24,3.55) -- (-.24,1.83) -- (policy.west);
\node[annotation,fill=white,inner sep=1pt,text=vsBlue] at (2.0,3.55) {initialize RL};
\end{tikzpicture}%
\endgroup%
}
\caption{\textbf{\methodname{}: data generation and two-stage policy training.} \textbf{(A)} Compiler search supplies verified schedules. Deterministic lifting preserves their executable edits in our Typed Schedule Edit Language (\TSEL{}); Kimi-K3 writes explanations from compiler-derived facts. Static and round-trip checks admit construction and refinement labels. \textbf{(B)} Weighted token supervision trains Seed-Coder-8B-Instruct to produce complete optimization graphs; $G_A$ denotes the optional prior graph for refinement. \textbf{(C)} Starting from SFT, GRPO updates the policy using numerical correctness and measured latency of sampled schedules.}
\label{fig:overview}
\vspace{-0.5cm}
\end{figure}

\methodname{} separates three responsibilities: the compiler defines and executes transformations, the policy chooses a complete schedule, and a verifier evaluates its correctness and performance. \Cref{fig:overview} separates training-data generation, supervised fine-tuning, and execution-grounded reinforcement learning.

\subsection{Optimization task and graph interface}
Apache TVM, an open-source deep learning compiler, constructs the unscheduled module $M_0$ in TensorIR, its loop-level intermediate representation, through \emph{lowering} (operators to loops and memory accesses) and \emph{task extraction} (isolating separately tunable operator groups). Rule-based analysis of its abstract syntax tree (AST) deterministically extracts structural features $F(M_0)$: loop extents, buffer shapes, reduction axes, and block structure. The policy receives $x=(d,M_0,F(M_0),h)$, with textual task description $d$ and compiler environment $h$ (target hardware and compiler configuration). Appendix~\ref{app:input-example} gives a concrete example.

A \emph{schedule} specifies compiler transformations such as loop tiling and GPU thread mapping. Its \emph{trace} $\tau$ records primitive calls, arguments, and sampling decisions:
\begin{equation}
 M_{\tau}=\operatorname{Replay}_h(M_0,\tau),\qquad
 K_{\tau}=\operatorname{Compile}_h(M_{\tau}).
 \label{eq:replay}
\end{equation}
Here $\operatorname{Replay}$ deterministically applies the trace, rejecting illegal transformations; $\operatorname{Compile}$ generates executable kernel $K_{\tau}$ from transformed module $M_{\tau}$. The subscript $h$ marks environment dependence. We seek a trace whose kernel has low measured latency $L_h(K_{\tau})$ on the target and passes a numerical comparison with the original computation. Refinement augments $x$ with an existing candidate graph $G_A$ and a summary of its schedule. The intended output in both modes specifies a complete trace rooted at $M_0$. A refined trace replaces $\tau_A$; it is not appended to the already transformed program.

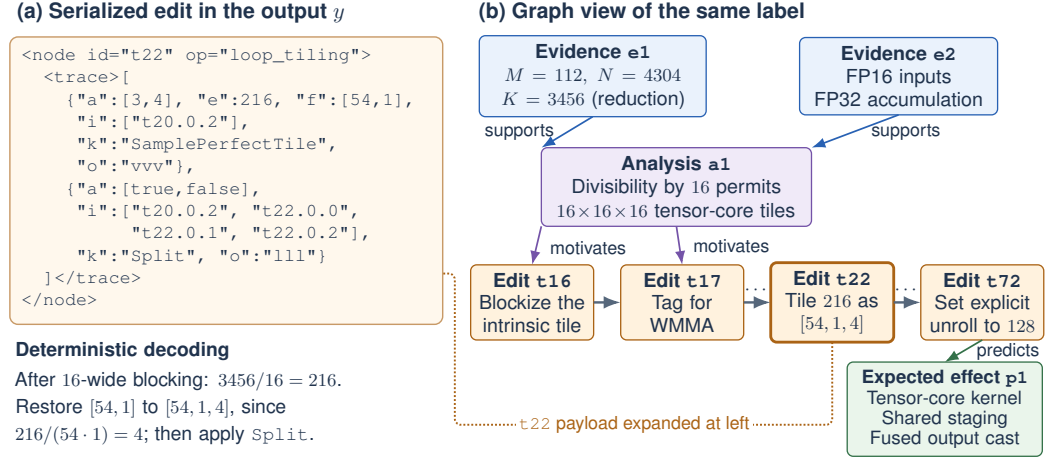
\begin{figure}[htbp]
\vspace{-0.3cm}
\centering
\resizebox{\linewidth}{!}{
\begingroup%
\definecolor{geInk}{HTML}{20334D}%
\definecolor{geMuted}{HTML}{56677B}%
\definecolor{geRule}{HTML}{CBD5E1}%
\definecolor{geBlue}{HTML}{2861B3}%
\definecolor{geBlueBg}{HTML}{EAF2FC}%
\definecolor{gePurple}{HTML}{7453A8}%
\definecolor{gePurpleBg}{HTML}{F0EAF8}%
\definecolor{geAmber}{HTML}{AA6315}%
\definecolor{geAmberBg}{HTML}{FFF2DE}%
\definecolor{geGreen}{HTML}{347149}%
\definecolor{geGreenBg}{HTML}{EAF5EB}%
\begin{tikzpicture}[
  x=1cm,y=1cm,>=Latex,text=geInk,
  font=\sffamily\fontsize{9.5}{11.5}\selectfont,
  box/.style={rounded corners=3pt,line width=.65pt,align=center,
    inner sep=4pt,outer sep=0pt},
  evidence/.style={box,draw=geBlue,fill=geBlueBg},
  analysis/.style={box,draw=gePurple,fill=gePurpleBg},
  edit/.style={box,draw=geAmber,fill=geAmberBg,text width=1.85cm,
    minimum height=1.3cm,font=\sffamily\fontsize{9.5}{11}\selectfont},
  effect/.style={box,draw=geGreen,fill=geGreenBg},
  semantic/.style={->,line width=.85pt},
  execution/.style={->,draw=geMuted,line width=1.2pt},
  edge label/.style={font=\sffamily\fontsize{8.8}{10}\selectfont,
    fill=white,inner sep=1.5pt},
  heading/.style={anchor=west,font=\sffamily\bfseries\fontsize{10.5}{12}\selectfont},
  note/.style={font=\sffamily\fontsize{9}{10.5}\selectfont,text=geMuted}
]
\path[use as bounding box] (0,0) rectangle (18.3,8.0);
\node[heading] at (.05,7.72) {(a) Serialized edit in the output $y$};
\node[heading] at (8.05,7.72) {(b) Graph view of the same label};

\path[draw=geAmber!65,fill=geAmberBg!55,rounded corners=4pt,line width=.7pt]
  (.05,2.32) rectangle (7.55,7.3);
\node[anchor=north west,inner sep=0pt] at (.24,7.18) {\begin{minipage}{7.15cm}
  \lstinputlisting[basicstyle=\ttfamily\fontsize{9}{11}\selectfont,
    frame=none,breaklines=false,columns=fullflexible,keepspaces=true,
    aboveskip=0pt,belowskip=0pt,showstringspaces=false]{figures/graph_example_t22.xml}
\end{minipage}};
\node[anchor=north west,align=left,text width=7.3cm,inner sep=0pt] at (.14,2.02) {
  \textbf{Deterministic decoding}\\[3pt]
  After $16$-wide blocking: $3456/16=216$.\\[2pt]
  Restore $[54,1]$ to $[54,1,4]$, since\\[2pt]
  $216/(54\cdot1)=4$; then apply \texttt{Split}.
};

\node[evidence,text width=3.65cm,minimum height=1.2cm] (e1) at (10.15,6.62) {
  \textbf{Evidence \texttt{e1}}\\
  $M=112,\ N=4304$\\
  $K=3456$ (reduction)};
\node[evidence,text width=3.15cm,minimum height=1.2cm] (e2) at (15.45,6.62) {
  \textbf{Evidence \texttt{e2}}\\
  FP16 inputs\\FP32 accumulation};
\node[analysis,text width=4.4cm,minimum height=1.2cm] (a1) at (11.6,4.7) {
  \textbf{Analysis \texttt{a1}}\\
  Divisibility by $16$ permits\\
  $16\!\times\!16\!\times\!16$ tensor-core tiles};
\draw[semantic,draw=geBlue] (e1.south) -- node[edge label,left=5pt,fill=none,pos=.48] {supports} (a1.north west);
\draw[semantic,draw=geBlue] (e2.south) -- node[edge label,right=7pt,fill=none,pos=.52] {supports} (a1.north east);

\node[edit] (t16) at (9.1,2.7) {
  \textbf{Edit \texttt{t16}}\\Blockize the\\intrinsic tile};
\node[edit] (t17) at (11.7,2.7) {
  \textbf{Edit \texttt{t17}}\\Tag for\\WMMA};
\node[edit,line width=1.4pt] (t22) at (14.3,2.7) {
  \textbf{Edit \texttt{t22}}\\Tile $216$ as\\$[54,1,4]$};
\node[edit] (t72) at (16.9,2.7) {
  \textbf{Edit \texttt{t72}}\\Set explicit\\unroll to $128$};
\draw[semantic,draw=gePurple] (a1.south west) -- node[edge label,right=5pt,fill=none,pos=.55] {motivates} (t16.north);
\draw[semantic,draw=gePurple] (a1.south) -- node[edge label,right=5pt,fill=none,pos=.5] {motivates} (t17.north);
\draw[execution] (t16.east) -- (t17.west);
\draw[execution] (t17.east) -- node[edge label,above=2pt,fill=none] {$\cdots$} (t22.west);
\draw[execution] (t22.east) -- node[edge label,above=2pt,fill=none] {$\cdots$} (t72.west);

\node[effect,text width=3.1cm,minimum height=1.25cm,
  font=\sffamily\fontsize{9}{10}\selectfont] (p1) at (16.25,.85) {
  \textbf{Expected effect \texttt{p1}}\\
  Tensor-core kernel\\Shared staging\\Fused output cast};
\draw[semantic,draw=geGreen] (t72.south) -- node[edge label,right=5pt,fill=none,pos=.52] {predicts} (p1.north);

\draw[draw=geAmber,densely dotted,line width=.95pt,rounded corners=3pt]
  (t22.south) -- (14.3,.58) -- (7.76,.58) -- (7.76,3.2) -- (7.55,3.2);
\node[edge label,text=geAmber] at (10.87,.58) {\texttt{t22} payload expanded at left};
\end{tikzpicture}%
\endgroup%
}
\caption{\textbf{A concrete optimization graph.} The payload fragment and graph projection come from the same FP16 projection training label. Node text is shortened; ellipses omit other edits. Colored edges $E$ explain choices; gray arrows show replay order $P$, and the dotted callout expands \code{t22}. WMMA denotes tensor-core operations. The expected effect \code{p1} concerns the whole schedule.}
\label{fig:graph-example}
\vspace{-0.7cm}
\end{figure}

The policy output $y=(y_1,\ldots,y_T)$ is the token sequence serializing a complete optimization graph $G=(V,E,P)$. Here $T$ is the output length, $y_t$ is its $t$-th token, and $y_{<t}$ is the preceding output prefix. The graph's nodes $V$ have four roles: evidence quotes facts from the input; analysis relates these facts to optimization opportunities; edits specify compiler actions in our \emph{Typed Schedule Edit Language} (\TSEL{}); and expected effects describe anticipated consequences. Typed edges $E$ connect evidence to analysis (\code{supports}), analysis to edits (\code{motivates}), and edits to effects (\code{predicts}). The payload $P$ orders edit groups and the instructions inside them. Replay follows this executable order, not a traversal of the explanatory edges. This separation allows exact execution and semantic explanation to be checked independently. \Cref{fig:graph-example} illustrates this separation with an actual projection training label: an edit's serialized payload determines its tiling factors, while $E$ provides explanatory links and $P$ fixes replay order.

\subsection{Compact and lossless schedule edits}
\TSEL{} represents each semantic edit as an ordered fragment of the TVM trace. For example, a tiling edit combines a sampling instruction with the split that consumes its factors; a memory-hierarchy edit introduces a cache stage. Grouping makes these actions explicit without replacing their executable parameters with prose.

\TSEL{} retains instruction kinds, typed inputs and outputs, attributes, sampling decisions, and the postprocessing boundary. Instead of global temporary-variable counters, a reference identifies its producer by \emph{edit group, local instruction, and output slot}. Index maps use a versioned, typed expression DAG, with shared maps stored once in a pool. The decoder reconstructs global handle numbering and canonical compiler objects. For every admitted source trace, the representation satisfies

\begin{equation}
 \operatorname{Canon}_h\!\left(\operatorname{Lower}_h(\operatorname{Lift}_h(\tau))\right)
 =\operatorname{Canon}_h(\tau).
 \label{eq:roundtrip}
\end{equation}
This round-trip condition preserves executable information while eliminating redundant serialization. It is checked under the recorded compiler and codec versions; it is not a claim of equivalence across arbitrary TVM releases.

The codec can also derive a redundant final tile factor. For loop extent $e$ and a factorization of arity $n$, it stores $e$ and $f_1,\ldots,f_{n-1}$ and reconstructs
\begin{equation}
 f_n=\frac{e}{\prod_{j=1}^{n-1}f_j},
 \label{eq:factor}
\end{equation}
rejecting nonintegral or otherwise illegal results. Reference resolution and this arithmetic check do not replace schedule replay: an edit must still be applicable to the program state created by its predecessors. \Cref{fig:graph-example} shows a serialized instruction fragment from a training label. Appendix~\ref{app:representation} provides the corresponding policy input and further details of the representation.

\subsection{Supervision from verified search}
Compiler search provides records $(M_0,\tau,M_{\tau},L)$ together with AST features and verification outcomes. We retain candidates that replay, compile, pass numerical checks, and have stable measured latency. Construction selects a fast candidate for each input. Refinement pairs a slower verified candidate with a faster candidate of the same computation, requiring the measured gap to exceed the prescribed reproducibility margin. Both endpoints retain their own complete traces; pairing does not assume that one trace is a valid incremental suffix of the other.

Label generation is split between deterministic tooling and a teacher model. The tooling lifts the chosen trace, derives facts about its effects on program structure, and fixes the exact edit payload. The teacher receives these facts and writes evidence, analyses, edit summaries, and expected effects. A static verifier checks cited sources, arithmetic derivations, applicability statements, and their consistency with the executable actions. Thus, an imprecise teacher explanation cannot silently change the transformation being taught.

The student receives only information available for its task: the initial program and features, and, for refinement, the supplied prior schedule and its summary. Target-side IR and measured target latency are kept out of the student prompt. The implemented refinement interface also includes an improvement-band request. Explanations provide supervised descriptions of schedule choices, not certificates that a particular edit caused a measured speedup.

\subsection{Supervised and execution-grounded policy learning}
SFT minimizes weighted autoregressive negative log-likelihood over graph completions:
\begin{equation}
 \mathcal L_{\mathrm{SFT}}
 =-\frac{\sum_{(x,y)}\sum_t w_t\log\pi_{\theta}(y_t\mid x,y_{<t})}
 {\sum_{(x,y)}\sum_t w_t}.
 \label{eq:sft}
\end{equation}
Consistent with prior SFT work on nonuniform token supervision \citep{kim2025sftgo,li2026later}, we use fixed weights $w_t=1$ for evidence, analysis, and expected-effect node spans; $w_t=4$ for sampled-decision payload spans, such as tiling-factor values, which directly govern schedule validity and kernel performance; and $w_t=2$ for all remaining output tokens, including trace structure and the end-of-sequence token. These spans are annotated deterministically before tokenization. A token crossing span boundaries receives the minimum weight of the characters it covers. Prompt tokens are masked out; in refinement dialogues, the previous assistant graph is context and only the final answer contributes to the loss. Thus ordinary executable content and sampled decisions receive twice and four times the weight of explanation-node content, respectively.

RL samples candidate graphs and executes their recovered traces. The final verification outcome determines the staged reward:
\begin{equation}
 R_{\mathrm{stage}}=
 \begin{cases}
 0, & \text{worker crash or unparseable payload},\\
 0.05, & \text{parsed, but validation or lowering fails},\\
 0.05+0.05\min(\max(f_{\mathrm{div}},p),0.99), & \text{lowered, but replay fails},\\
 0.15, & \text{replayed, but compilation fails},\\
 0.35, & \text{compiled, but numerically incorrect},\\
 1+r_{\mathrm{lat}}, & \text{numerically correct}.
 \end{cases}
 \label{eq:stage-reward}
\end{equation}
Here $f_{\mathrm{div}}$ is the emitted-tile divisibility fraction, $p$ is the replayed-instruction fraction, and $r_{\mathrm{lat}}$ is the latency bonus for a numerically correct candidate. We define numerical correctness as successful trace replay and compilation followed by output agreement with the LLVM-compiled original TensorIR using one randomly generated test case per workload. Intermediate rewards distinguish progress through parsing, replay, and compilation, providing learning signals for partially successful proposals, while latency bonuses require numerical correctness, consistent with previous work \citep{dai2026cudaagent}.

To compare latency gains across workloads, let $L_0$ be the default-scheduled latency and $a<L_0$ a fixed reference anchor. We measure kernel latency as the median of five synchronized timing repeats after warmup, and calibrate task-specific noise bands by benchmarking identical schedules in three separate evaluator runs, each launched in a fresh process. Define the normalized utility
\begin{equation}
 u(L)=\frac{\log(L_0/L)}{\log(L_0/a)},\qquad
 Q_m(u)=q_m\left\lfloor u/q_m\right\rfloor,
 \label{eq:utility}
\end{equation}
where $q_m$ is a noise-dependent quantum for task mode $m$. Quantization prevents small timing fluctuations from always producing a distinct reward. The latency bonus in \cref{eq:stage-reward} is
\begin{equation}
 r_{\mathrm{lat}}=\alpha_m b_m(L)\clip\!\left(Q_m(u(L)),0,u_{\max}\right),
 \label{eq:reward}
\end{equation}
Here $\alpha_m$ is the mode-specific latency-reward coefficient: $\alpha_{\mathrm{main}}=0.30$ for construction and $\alpha_{\mathrm{refine}}=0.50$ for refinement. The cap is $u_{\max}=1.2$, and $b_m$ requires a sufficiently large improvement over the default schedule for construction and over the supplied parent for refinement. The latency coefficient is bounded so that ordinary correct candidates outrank incorrect ones.

We optimize this reward with group-relative policy optimization \citep{shao2024deepseekmath}. Advantages compare sampled candidates for the same input, and zero-variance groups are excluded from updates. The executable payload determines the reward independently of its explanatory text. Partial credit reflects progress through the execution pipeline, while the latency bonus requires numerical correctness and a measured improvement satisfying the prescribed gates. A candidate therefore cannot earn performance reward merely by producing a persuasive explanation or claiming a speedup: its recovered trace must produce a kernel whose measured behavior earns that reward. This separation provides an explicit safeguard against reward hacking through unsupported optimization claims. Details on reinforcement learning optimization policy appear in Appendix~\ref{app:reward}.

\FloatBarrier
\section{Results}
\label{sec:results}

Following the experimental setup (Section~\ref{sec:protocol}), we examine four questions. Section~\ref{sec:sft-results} asks whether search-derived supervision can teach an 8B model to construct and refine optimizations, showing high correctness and substantial kernel speedups on held-out tasks. Section~\ref{sec:rl-results} tests whether reinforcement learning feedback improves this policy, finding further kernel speedups while maintaining high correctness. Section~\ref{sec:api-results} examines whether the trained model can outperform frontier baselines, finding that it leads all six. Section~\ref{sec:budget} compares its optimization cost with TVM search, and the model produces faster kernels with substantially less optimization time.

\subsection{Experimental setup}
\label{sec:protocol}
\paragraph{Data and training.}
Our workloads span matrix multiplication, attention, normalization and reductions, mixture-of-experts operations, convolution, and quantization. We fine-tune Seed-Coder-8B-Instruct \citep{seed2025seedcoder} for two epochs on 59,040 construction and 47,232 refinement examples, with explanation labels generated by Kimi-K3 \citep{kimi2026k3}, then apply execution-grounded RL. Splits hold out entire workload keys and check task names and initial-IR hashes. SFT is selected using validation; the reported RL checkpoint is selected using a post-hoc validation-latency criterion.

\paragraph{Evaluation and baselines.}
We evaluate on NVIDIA H100 using 116 held-out construction tasks and 194 refinement prompts spanning 39 tasks. We separate examples by workload identity and verify zero overlap between the sealed evaluation tasks and the SFT training, validation, and RL training pools using both workload identities and recorded structural hashes of the initial TensorIR modules. A task is a statically shaped TensorIR computation; a prompt is one model input. Construction supplies one prompt per task, whereas refinement supplies multiple prompts for some tasks, with different parent schedules or refinement requests. Each model generates eight outputs per prompt, totaling 928 construction outputs and 1,552 refinement outputs. Recovered programs are compiled, checked numerically, and timed under the recorded protocol. Compiler references are TVM's default GPU schedule ($L_0$) and its archived MetaSchedule search results. We compare SFT and SFT+RL with GPT-6 Astra \citep{openai2026astra}, Opus 5 \citep{anthropic2026opus5}, Seed-Coder-8B-Instruct, Kimi-K3, GLM-5.3 \citep{zai2026glm53}, and Qwen3.8-27B \citep{qwen2026qwen38}. For construction, external baselines receive PyTorch reference code and generate complete, self-contained TVMScript TensorIR kernels. This interface avoids requiring them to emit our custom optimization-graph representation. For refinement, they generate complete optimization graphs from workload--parent inputs.

\paragraph{Metrics.}
Correct@1 is the $k{=}1$ case of pass@$k$~\citep{chen2021codex} with numerical correctness as success. We define construction $\mathrm{fast}_p$ as the percentage of all generated outputs that are numerically correct and achieve speedup $L_0/L>p$, where $L$ is the candidate latency and $L_0$ is the task's default-scheduled latency; \Cref{tab:api-test2} reports $\mathrm{fast}_2$ and $\mathrm{fast}_{10}$. Refinement performance columns instead require correctness and improvement over the supplied parent, graph A: either beyond the task's recorded noise band or by at least $2\times$, using the parent's latency $L_A$ as the reference. Untimed outputs receive no performance credit. Paired SFT--RL speedup-win counts use a separate population: prompts for which both models' fixed first outputs are correct and timed.

\begin{table}[!t]
\centering\footnotesize\setlength{\tabcolsep}{1pt}
\begin{tabular*}{\linewidth}{@{\extracolsep{\fill}}lrrr@{\hspace{9pt}}rrr@{}}
\toprule
& \multicolumn{3}{c}{\textbf{Construction (\%)}} & \multicolumn{3}{c}{\textbf{Refinement (\%)}} \\
\cmidrule(lr){2-4}\cmidrule(l){5-7}
Model & Correct@1 & $\mathrm{fast}_2$ & $\mathrm{fast}_{10}$ & Correct@1 & \begin{tabular}[c]{@{}c@{}}Correct and\\improves \\graph A\end{tabular} & \begin{tabular}[c]{@{}c@{}}Correct and\\$\geq2\times$ over\\ graph A\end{tabular} \\
\midrule
GPT-6 Astra & 9.38 & 7.65 & 6.03 & 45.55 & 36.47 & 17.33 \\
Opus 5 & 64.01 & 58.30 & 36.96 & 45.81 & 31.70 & 14.24 \\
Kimi-K3 & 34.91 & 25.43 & 14.55 & 37.82 & 19.59 & 4.70 \\
GLM-5.3 (Z.ai) & 15.73 & 11.75 & 7.00 & 18.56 & 8.25 & 1.80 \\
Qwen3.8-27B & 0.11 & 0.00 & 0.00 & 19.59 & 4.57 & 1.48 \\
Seed-Coder-8B-Instruct & 0.00 & 0.00 & 0.00 & 8.12 & 4.51 & 1.22 \\
\midrule[1pt]
SFT (ours) & 89.33 & 84.16 & 64.98 & 91.95 & 68.56 & 37.24 \\
SFT+RL (ours) & 91.27 & 89.33 & 77.69 & 92.01 & 87.05 & 59.28 \\
\bottomrule
\end{tabular*}

\caption{\textbf{Per-output correctness and optimization success.} Results cover 116 construction tasks and 194 refinement prompts spanning 39 tasks.}
\label{tab:api-test2}
\vspace{-0.5cm}
\end{table}

\subsection{SFT transfers compiler-search experience to an 8B model}
\label{sec:sft-results}
\textbf{Search-derived supervision teaches the small model to generate complete, effective optimizations on held-out tasks.} SFT achieves Correct@1 of $89.33\%$ for construction and $91.95\%$ for refinement (\cref{tab:api-test2}). Each accepted proposal specifies a complete transformation script that replays, compiles, and passes the numerical checks. The performance columns show that this capability includes substantial optimization: SFT achieves construction $\mathrm{fast}_{10}$ of $64.98\%$, while $68.56\%$ of refinement outputs correctly improve the supplied graph A beyond the recorded noise band. These results support both training objectives from the introduction: learning to construct an optimization from an unoptimized kernel and learning to improve an existing optimization. SFT thus supplies a policy that can already produce useful complete scripts, providing the starting point for learning from their outcomes.

\FloatBarrier
\subsection{RL improves schedule quality through execution feedback}
\label{sec:rl-results}
\textbf{Execution feedback directs an already capable policy toward faster kernels.} At the selected RL checkpoint, construction $\mathrm{fast}_{10}$ rises from $64.98\%$ to $77.69\%$. For refinement, the fraction that are correct and at least twice as fast as graph A rises from $37.24\%$ to $59.28\%$ (\cref{tab:api-test2}). Correctness remains high, increasing from $89.33\%$ to $91.27\%$ for construction and from $91.95\%$ to $92.01\%$ for refinement. The larger changes in performance-success rates show where RL contributes most: it strengthens the model's ability to propose a useful optimization once SFT has established executable generation.

\textbf{The improvement is visible in the fixed first proposal.} Among prompts with correct, timed first outputs from both models, RL produces faster kernels on 88 of 102 construction tasks and 145 of 170 refinement prompts. Together with the threshold-success rates, this supports the introduction's hypothesis that feedback on the model's own complete proposals can improve the optimization policy learned from search-derived demonstrations.

\subsection{\methodname outperforms frontier baselines}
\label{sec:api-results}
\textbf{The trained 8B model leads all six reported metrics across construction and refinement.} \Cref{tab:api-test2} compares it with six external baselines, including GPT-6 Astra and Opus 5. The comparison also includes much larger open models: Kimi-K3 has 2.8T total parameters, with 104B activated per token \citep{kimi2026k3}, and GLM-5.3 has approximately 753B total parameters \citep{zai2026glm53}.

\begin{figure}[t]
\centering
\includegraphics[width=\linewidth]{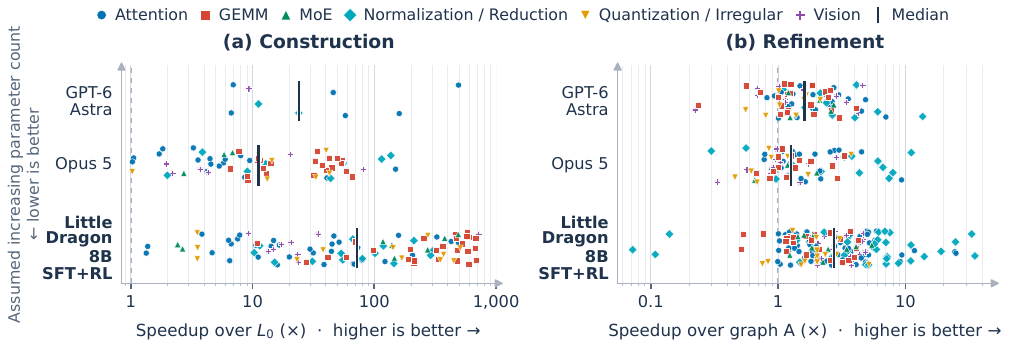}
\caption{\textbf{First output kernel speedups.} Our \textsc{Little Dragon}\xspace achieves better speedups for both the construction and the refinement tasks across multiple kernel famiilies.}
\label{fig:frontier-speedups}
\vspace{-0.5cm}
\end{figure}

\paragraph{Training a small model yields a large correctness advantage.}
As shown in \Cref{tab:api-test2}, for construction, SFT+RL reaches $91.27\%$ Correct@1, compared with $64.01\%$ for the strongest external baseline, Opus 5---a $27.26$ percentage-point advantage. For refinement, the corresponding rates are $92.01\%$ and $45.81\%$, a $46.20$ percentage-point advantage. SFT alone already exceeds every external baseline in all six columns; RL extends that advantage. This progression connects the comparison to the training method: search-derived supervision establishes strong construction and refinement performance, and execution feedback further improves the trained policy.

\paragraph{\textsc{Little Dragon}\xspace improves kernel speedup substantially with a much smaller model size.}
For construction, SFT+RL achieves $\mathrm{fast}_{10}$ of $77.69\%$, compared with $36.96\%$ for Opus 5, the strongest external model on this metric. For refinement, $59.28\%$ of SFT+RL outputs are correct and at least twice as fast as their supplied graph A, compared with $17.33\%$ for GPT-6 Astra, the strongest external model on that metric. These comparisons retain failures in the denominator, measuring how often one attempt yields a correct kernel with a substantial performance gain. They support the paper's central claim: training on compiler-search experience and execution outcomes enables an 8B model to outperform frontier baselines, including models with hundreds of billions to trillions of parameters, on the reported optimization tasks. \Cref{fig:frontier-speedups} complements these success rates with speedup distributions for correct, timed first outputs on the plotted subsets. The trained policy's median speedups are $71.8\times$ over $L_0$ for construction and $2.74\times$ over graph A for refinement, compared with $11.2\times$/$1.26\times$ for Opus~5 and $24.0\times$/$1.61\times$ for GPT-6 Astra.

\begin{figure}[t]
\vspace{-4mm}
\centering
\includegraphics[width=\linewidth]{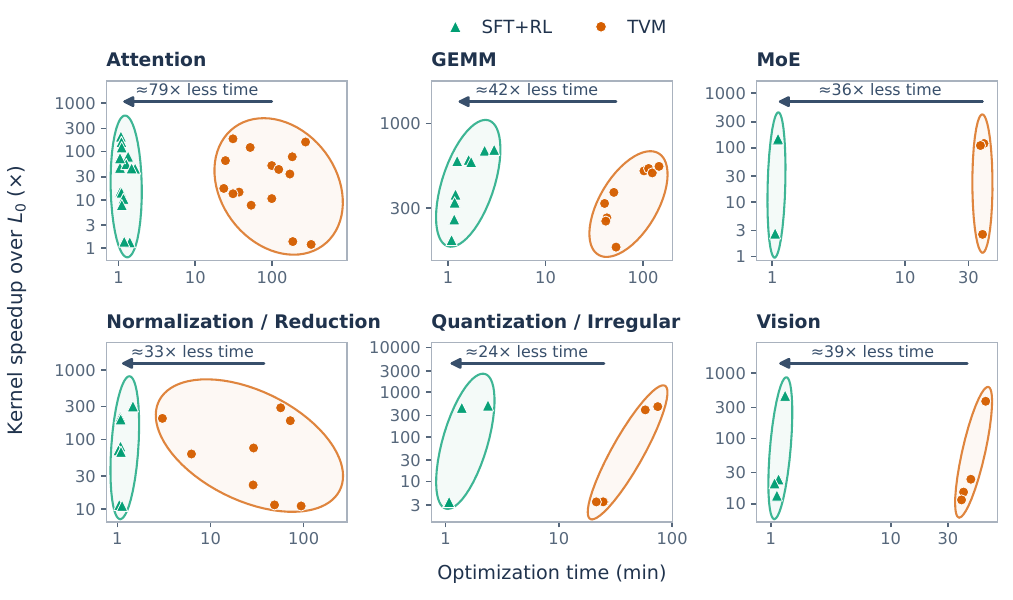}
\vspace{-0.8cm}
\caption{\textbf{Kernel speedup and optimization cost comparisons} by family between first SFT+RL output and TVM MetaSchedule search run.}
\label{fig:selected-workloads}
\vspace{-0.8cm}
\end{figure}

\subsection{Comparison with TVM search}
\label{sec:budget}
\textbf{\textsc{Little Dragon}\xspace achieves better kernels with substantially less per-task optimization
time.} \Cref{fig:selected-workloads} compares optimization time and kernel quality across six workload families between first SFT+RL output and TVM MetaSchedule search run. Under the plotted timing convention, SFT+RL produces these faster kernels with approximately $39\times$ less optimization time, measured by the median paired TVM-to-model time ratio. The family medians range from $24\times$ to $79\times$: $79\times$ for Attention, $42\times$ for GEMM, $36\times$ for MoE, $33\times$ for Normalization/Reduction, $24\times$ for Quantization/Irregular, and $39\times$ for Vision. Kernel speedups use each task's own default schedule, $L_0$. In the Attention panel, dense MLA decode reaches $44.45\times$ speedup over $L_0$ with SFT+RL versus $34.25\times$ with TVM, at plotted optimization times of $1.64$ versus $170.94$ minutes---approximately $104\times$ less time. In the GEMM panel, FP8 matrix multiplication reaches $585.33\times$ versus $373.98\times$ speedup, using $1.24$ versus $50.22$ minutes, or approximately $41\times$ less time. In the Normalization/Reduction panel, the fused softmax task in a matmul--GELU--softmax workload reaches $70.49\times$ versus $22.18\times$ speedup, using $1.03$ versus $28.66$ minutes, or approximately $28\times$ less time. These examples show how compiler-search experience can be reused through training: the 8B policy produces a complete
schedule in a single proposal, reducing the per-task cost of obtaining an optimized kernel.

\vspace{-0.15cm}
\section{Related Work}
\vspace{-0.1cm}
\paragraph{Compiler and LLM-guided search.}
TVM separates computation from scheduling; Ansor and MetaSchedule search transformations~\citep{chen2018tvm,zheng2020ansor,shao2022metaschedule}, while AutoTVM learns cost models to guide measurement~\citep{chen2018learning}. Reasoning Compiler and LiteCoOp use LLM-guided search trees~\citep{tang2025reasoningcompiler,tang2026litecoop}; AccelOpt, AlphaEvolve, and KernelEvolve combine model proposals with execution feedback and search or evolution~\citep{zhang2026accelopt,novikov2025alphaevolve,liao2025kernelevolve}. These LLM-guided methods keep the proposing models fixed.
\vspace{-0.1cm}
\paragraph{Learning optimizations and kernels.}
Cummins et al.\ and LLM Compiler train models for compiler optimization~\citep{cummins2023large,cummins2024llmcompiler}; Compiler-R1 and ECCO incorporate performance-rewarded training~\citep{pan2025compilerr1,pan2026ecco}. TLM and Mehmood et al.\ learn schedule generation~\citep{zhai2024enabling,mehmood2026tlm}. Kevin, AutoTriton, Dr.\ Kernel, and CUDA Agent train kernel generators with execution feedback~\citep{baronio2025kevin,li2025autotriton,liu2026drkernel,dai2026cudaagent}, while TritonRL adds safeguards against reward hacking~\citep{woo2025tritonrl}.
\vspace{-0.1cm}
\paragraph{Representations and our approach.}
Learned edit representations transfer program changes~\citep{yin2019learning}, while CAKE exposes typed schedules and evolves its compiler harness~\citep{ye2026cake}. Our policy learns complete construction and refinement schedules through search-derived supervision and execution-grounded RL. Our typed edit language reconstructs exact compiler traces, with explanations grounded in compiler-derived facts. Replay and compilation transform the existing computation into a kernel, which undergoes numerical validation.

\vspace{-0.15cm}
\section{Conclusion}
\vspace{-0.1cm}
We introduced \methodname{} for training a small language model to construct and refine complete GPU kernel optimization scripts. Our compact representation enables exact reconstruction of compiler traces. SFT teaches construction and refinement tasks from well-performing verified compiler search examples; RL improves schedule choices through correctness and latency feedback. The trained 8B model achieves Correct@1 of $91.27\%$ for construction and $92.01\%$ for refinement, outperforming six baselines across all reported correctness and performance-success metrics. These results establish compiler search experience and execution feedback as effective training signals for a policy that coordinates compiler optimization scripts in one complete proposal.

\label{lastmainpage}
\bibliography{references}
\bibliographystyle{iclr2026_conference}
\clearpage
\appendix

\section{Representation and training-label example}
\label{app:representation}

\subsection{A concrete policy input}
\label{app:input-example}
The construction input $x=(d,M_0,F(M_0),h)$ below comes from the projection training example used in \Cref{fig:graph-example}. It computes $Y=\operatorname{cast}_{\mathrm{FP16}}(XW^\top)$. The task description and feature record are abbreviated for readability. TensorIR identifiers are shortened, with $X$, $W$, $C$, and $Y$ denoting the buffers.

\paragraph{Task description $d$.}
``Optimize the extracted TensorIR task \code{fused\_NT\_matmul\_cast}: an NT projection with 112 rows, input width 3456, and output width 4304, for lower latency on NVIDIA H100. The generated schedule must preserve numerical correctness.''

\paragraph{Initial module $M_0$.}
The unscheduled module contains a matrix-product reduction followed by an output cast. Its serial loops specify the computation before tiling, thread binding, or tensorization:
\begin{lstlisting}[language=Python]
from tvm.script import ir as I, tirx as T

@I.ir_module
class Module:
    @T.prim_func(s_tir=True)
    def main(X: T.Buffer((T.int64(112), T.int64(3456)), "float16"),
             W: T.Buffer((T.int64(4304), T.int64(3456)), "float16"),
             Y: T.Buffer((T.int64(112), T.int64(4304)), "float16")):
        T.func_attr({"tirx.noalias": True})
        C = T.sblock_alloc_buffer((T.int64(112), T.int64(4304)), "float32")
        for i, j, k in T.grid(T.int64(112), T.int64(4304), T.int64(3456)):
            with T.sblock("NT_matmul"):
                vi, vj, vk = T.axis.remap("SSR", [i, j, k])
                T.reads(X[vi, vk], W[vj, vk])
                T.writes(C[vi, vj])
                with T.init():
                    C[vi, vj] = T.float32(0.0)
                C[vi, vj] = C[vi, vj] + (
                    T.Cast("float32", X[vi, vk]) * T.Cast("float32", W[vj, vk]))
        for i, j in T.grid(T.int64(112), T.int64(4304)):
            with T.sblock("compute"):
                vi, vj = T.axis.remap("SS", [i, j])
                T.reads(C[vi, vj])
                T.writes(Y[vi, vj])
                Y[vi, vj] = T.Cast("float16", C[vi, vj])
\end{lstlisting}

\paragraph{Structural features $F(M_0)$.}
Deterministic AST analysis records one function, five serial loops, and no thread bindings. A subset of the actual feature fields is:
\begin{lstlisting}
{"primfunc_count": 1, "loop_kind_counts": {"serial": 5},
 "thread_binding_counts": {}, "storage_scope_counts": {"global": 4}}
\end{lstlisting}
The same record lists the buffers' shapes and dtypes, the reduction block's two spatial axes of extents 112 and 4304 and reduction axis of extent 3456 (\code{SSR}), and the cast block's two spatial axes (\code{SS}). Its read/write regions show that the reduction reads $X$ and $W$ and writes the intermediate $C$, which the cast reads to produce $Y$. These facts are extracted from $M_0$ before a schedule is proposed.

\paragraph{Compiler and target environment $h$.}
The recorded environment uses TVM \code{0.26.dev0}, and an NVIDIA H100 80GB HBM3 CUDA target with architecture \code{sm\_90}. The target configuration specifies a warp size of 32, at most 1024 threads per block, and a shared-memory-per-block limit of 49,152 bytes.

\subsection{The two layers of an optimization graph}
The executable payload and the explanation layer have different contracts. The payload must reconstruct one complete pinned TVM trace, including its sampling choices and postprocessing. Its validity is tested by the compiler and empirical execution. The explanation layer identifies evidence, proposes reasons for choices, and links those reasons to edits. During label generation, teacher explanations are checked against facts from the supplied input and the actual payload; these checks do not provide a formal proof of performance causality.

The serialization uses an XML graph containing JSON trace fragments. The exact order of \code{trace\_sequence} nodes and their instruction lists determines execution. The principal instruction fields are \code{k} (kind), \code{i} (inputs), \code{a} (attributes), \code{o} (output type signature), and a sampled decision or its lossless derived-factor encoding. A reference such as \code{t22.0.2} denotes output 2 of local step 0 in edit group \code{t22}. Type signatures distinguish block, loop, and expression values. These local coordinates avoid asking the model to maintain a global output counter, but still require valid backward references.

\textbf{Losslessness scope.} Equation~\ref{eq:roundtrip} concerns the canonical TVM trace representation under the recorded codec/compiler version. It does not imply arbitrary historical Python whitespace reproduction, equivalence between compiler versions, or universal support for every possible TVM object. Accepted dataset traces are round-trip checked. Unsupported constructs must fail validation rather than be silently approximated. The lowering layer retains every instruction even when the explanatory description groups several instructions into one semantic edit.

\newpage
\section{RL policy optimization}
\label{app:reward}
For eight sampled completions, the trainer forms group-standardized advantages $A_i=(R_i-\overline R)/(s_R+\epsilon)$ and uses the PPO-style clipped likelihood-ratio objective \citep{schulman2017ppo} of the form
\begin{equation}
 J(\theta)=\mathbb E\left[\frac{1}{8}\sum_{i=1}^{8}\frac{1}{|y_i|}
 \sum_t\min\!\left(r_{i,t}(\theta)A_i,
 \clip(r_{i,t}(\theta),1-0.2,1+0.2)A_i\right)\right].
\end{equation}
$r_{i,t}$ is the new-policy to rollout-policy token probability ratio. There is no KL term. Dynamic sampling drops groups with zero reward variance, following the filtering idea in DAPO \citep{yu2025dapo}, and obtains additional groups from an oversampled pool (160 prompts for a retained batch of 128). This makes the retained training distribution policy-dependent. The 60,800 target rows are selected from an SFT-screened pool with a $60{:}40$ construction/refinement mixture, so training performance is not a random estimate of the entire workload universe.

\begin{figure}[ht]
\centering\includegraphics[width=.84\linewidth]{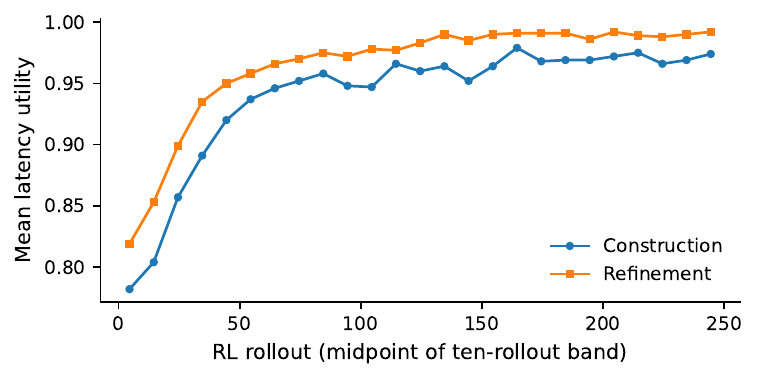}
\caption{\textbf{RL training utility.} Ten-rollout means, plotted at band midpoints. The series is measured on screened, dynamically retained training samples, not the sealed set.}
\end{figure}

\begin{figure}[ht!]
\centering\includegraphics[width=.84\linewidth]{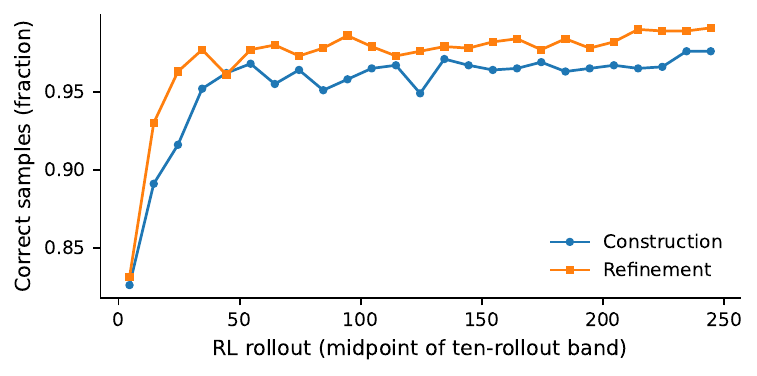}
\caption{\textbf{RL training correctness.} Ten-rollout sample-correctness summaries.}
\end{figure}

\FloatBarrier
\section{Code Repository}
\label{app:code}
Example scripts and workloads accompanying this paper are available at
\url{https://github.com/Anna-Bele/How-To-Train-Your-Little-Dragon}.

\end{document}